\documentclass[a4paper,fleqn]{cas-sc} 
\usepackage[authoryear,longnamesfirst]{natbib}
\usepackage{tikz}
\usetikzlibrary{positioning,arrows.meta,calc}

\ExplSyntaxOn
\cs_set:Npn \__first_footerline: { }
\ExplSyntaxOff

\def\tsc#1{\csdef{#1}{\textsc{\lowercase{#1}}\xspace}}
\tsc{WGM}
\tsc{QE}

\begin{document}
\let\WriteBookmarks\relax
\def\floatpagepagefraction{1}
\def\textpagefraction{.001}

\shorttitle{Task-Sensitive Geometry of Representation Transfer}

\shortauthors{Pham}  

\title[mode=title]{Task-Sensitive Geometry of Representation Transfer for Object Detection under Image Degradation}

\author[1]{Van Vung Pham}[orcid=0000-0001-9702-8904]
\ead{vung.pham@shsu.edu}

\affiliation[1]{organization={Computer Science Department, Sam Houston State University},
            city={Huntsville},
            state={Texas},
            country={USA}}


\begin{abstract}
Object detection under image degradation can benefit from clean-image supervision, but aggregate gains do not imply that transferred representation changes are uniformly useful. We study how clean task knowledge affects degraded-image representations and whether local responses to structured representation directions can be characterized geometrically. Using paired clean and Gaussian-degraded BDD100K images, we show that clean-teacher distillation improves observed detection accuracy while producing heterogeneous object-level transfer. We isolate a representation component complementary to direct clean-teacher alignment and map it into the distilled student space through an orthogonal bridge. Controlled interventions rescue 13.22\% of objects lost under the distilled representation, versus 4.30\% under norm-matched random perturbations, with very low harm on preserved objects. We introduce task-sensitive geometry, a gradient-derived channel-space geometry constructed from normalized detection-loss gradients. On a reserved cohort, mapped-complement orientation within this frozen geometry is positively associated with local intervention-response magnitude after controlling for intervention magnitude (partial Spearman $\rho=0.242$, 95\% CI $[0.108,0.359]$). The relationship replicates on independent data ($\rho=0.180$) and with RT-DETR-L ($\rho=0.227$), but not for the direct clean-teacher residual family, and it weakens for large interventions. Routing rules and specialized distillation objectives based on these signals do not yield statistically reliable gains over CLEAN-KD. These results support a local, direction-family-dependent task-sensitive geometry while showing that converting such structure into improved global training remains an open problem.
\end{abstract}




\maketitle

\section{Introduction}
\label{sec:introduction}

Object detection systems increasingly operate under imaging conditions that differ substantially from the clear, high-quality data on which modern detectors are commonly developed. Noise, low illumination, rain, fog, blur, and other acquisition degradations can suppress or distort visual evidence needed for localization and recognition, producing substantial performance degradation in downstream detectors \citep{Gupta2024WeatherDetection,Ogino2025ERUPYOLO}. This problem is especially relevant in autonomous and safety-critical vision, where datasets such as BDD100K \citep{Yu2020BDD100K} and ACDC \citep{Sakaridis2021ACDC} explicitly capture substantial variation in environmental and adverse visual conditions. Improving robustness under such conditions therefore requires more than recovering visually plausible images; it requires preserving or recovering information that is useful to the downstream task.

One approach is to exploit a higher-quality observation during training. When paired clean and degraded views of the same scene are available, a detector operating on the clean image can provide supervisory information to a detector that must ultimately operate on degraded input. Knowledge distillation provides a natural mechanism for such transfer \citep{Hinton2015KD}. In object detection, increasingly sophisticated methods have shown that the effectiveness of distillation depends strongly on which spatial regions, features, instances, scales, or prediction components are transferred \citep{Wang2019FineGrainedKD,Guo2021DecoupledFeatures,Dai2021GID,Yang2022FGD,Zhu2023ScaleKD,Wang2024CrossKD,Wang2024KDDETR}. These advances demonstrate that detector knowledge is not uniformly informative across the representation. However, most distillation methods ultimately optimize a teacher-derived target once the transferred components have been selected. They do not directly determine whether a particular feature-space displacement is locally beneficial to the downstream detection objective.

A related observation has emerged in task-driven image restoration. Improvements in perceptual image quality do not necessarily translate into improvements in recognition or detection, motivating restoration objectives that incorporate downstream task information \citep{Marathe2022RestoreXAI,Kim2024SR4IR,Kim2025EDTR,Chen2025UniRestore}. This distinction between visual fidelity and task utility is fundamental. The analogous issue arises inside a detector: a feature representation that is closer to a clean-teacher representation is not necessarily synonymous with a locally more useful representation for the downstream task. Clean-teacher supervision may improve detection on average while individual objects can be preserved, rescued, lost, or remain missed under transfer.

This work studies that representation-level question. We first establish clean-to-degraded knowledge transfer in a paired detection setting and then analyze the resulting transfer at the object level. The analysis reveals substantial heterogeneity: clean-teacher support is strongly associated with preservation and rescue, yet locally unfavorable transfer remains possible under an overall beneficial objective. We further isolate a representation component that is complementary to direct clean-teacher alignment and test it through controlled feature interventions. These experiments show that structured complementary directions can rescue objects lost under the distilled representation substantially more often than norm-matched random perturbations, without producing a corresponding increase in harm to already preserved objects.

\begin{figure}
\centering

\begin{tikzpicture}[
    font=\small,
    >=Latex,
    box/.style={
        draw,
        rounded corners=2pt,
        align=center,
        minimum height=1.15cm,
        text width=3.8cm,
        inner sep=4pt,
        line width=0.7pt
    },
    smallbox/.style={
        draw,
        rounded corners=2pt,
        align=center,
        minimum height=1.15cm,
        text width=2.5cm,
        inner sep=4pt,
        line width=0.7pt
    },
    arrow/.style={
        ->,
        line width=0.8pt
    },
    dashedarrow/.style={
        ->,
        dashed,
        line width=0.8pt
    }
]


\node[smallbox] (clean)
{Clean image\\
$I_c$};

\node[smallbox, below=1.0cm of clean] (degraded)
{Degraded image\\
$I_d$};


\node[smallbox, right=0.5cm of clean] (teacher)
{Clean teacher\\
$T$};

\node[smallbox, right=0.5cm of degraded] (student)
{Degraded student\\
$S$};

\draw[arrow] (clean) -- (teacher);
\draw[arrow] (degraded) -- (student);

\draw[arrow]
(teacher.south) --
node[left, font=\footnotesize] {CLEAN-KD}
(student.north);


\node[box, right=1.0cm of teacher] (heterogeneous)
{Object-level transfer\\[1mm]
PRES/LOST/RESC/MISS};

\draw[arrow]
(student.north east)
to[out=0,in=180]
(heterogeneous.west);


\node[box, below=1.0cm of heterogeneous] (complement)
{Mapped complement\\[1mm]
$c_i \rightarrow m_i$, 
$S+\alpha D_M$};

\draw[arrow]
(student.east) -- (complement.west);


\node[box, right=1.0cm of heterogeneous] (geometry)
{Task-sensitive geometry\\[1mm]
$g_i \rightarrow C_g$, 
$D_M \rightarrow Q_M$};

\draw[arrow]
(heterogeneous.east) -- (geometry.west);


\node[box, below=1.0cm of geometry] (validation)
{Prospective evaluation, 
$Q_M \leftrightarrow Y_M$};

\draw[arrow]
(complement.east) -- (validation.west);

\draw[arrow]
(geometry.south) -- (validation.north);


\node[
    font=\footnotesize\bfseries,
    anchor=base
] at ($($(clean.north)!0.5!(teacher.north)$)+(0,0.3)$)
{Paired clean--degraded transfer};

\node[
    font=\footnotesize\bfseries,
    anchor=base
] at ($(heterogeneous.north |- clean.north)+(0,0.3)$)
{Transfer heterogeneity};

\node[
    font=\footnotesize\bfseries,
    anchor=base
] at ($(geometry.north |- clean.north)+(0,0.3)$)
{Geometric characterization};

\end{tikzpicture}

\caption{
Overview of the proposed framework.
Paired clean and degraded observations establish clean-to-degraded
representation transfer through CLEAN-KD.
Object-level analysis reveals heterogeneous transfer outcomes,
while complementary representation interventions identify
task-relevant information beyond direct clean-teacher alignment.
Task-sensitive geometry characterizes candidate representation
directions using downstream task gradients and is evaluated
prospectively against measured intervention responses.
}
\label{fig:framework}

\end{figure}
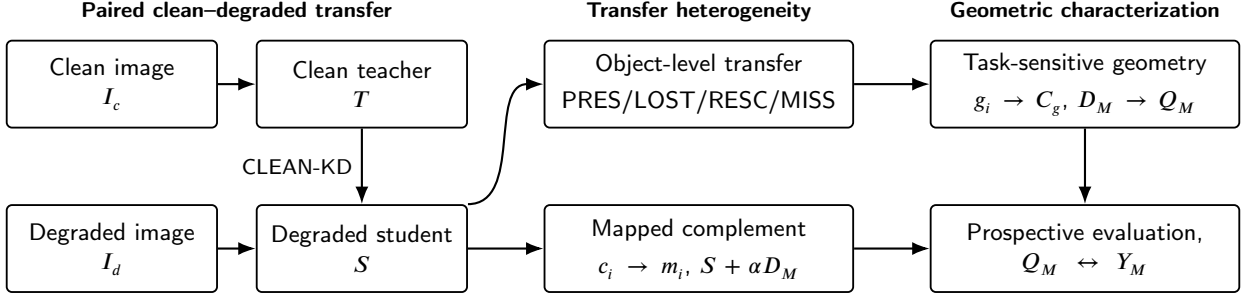

These observations motivate a geometric characterization of local task sensitivity. We introduce \emph{task-sensitive geometry} (TSG), constructed from normalized gradients of the downstream detection loss with respect to an internal feature representation. Rather than describing which feature directions are statistically common, TSG describes channel-space orientations to which the downstream task is repeatedly sensitive. A candidate representation displacement can then be characterized by its orientation within this frozen geometry. The central hypothesis is that, within a structured intervention family,
directions more strongly aligned with task-sensitive geometry should exhibit
systematically different downstream responses.
Figure~\ref{fig:framework} summarizes the overall framework, from paired
clean-to-degraded representation transfer and heterogeneous object-level
outcomes to complementary interventions, task-sensitive geometry, and
prospective response evaluation.

A critical aspect of the study is prospective separation between geometry construction and response evaluation. The task-sensitive geometry is frozen before intervention outcomes on the confirmation cohort are observed. Its mapped-complement orientation score is then tested against subsequently measured local task response while controlling for intervention magnitude. The resulting association is positive with a confidence interval entirely above zero, replicates on an independent fresh cohort, and is reproduced with RT-DETR-L. At the same time, the relationship weakens for large interventions and does not generalize to the direct clean-teacher residual direction family. These boundary results are important: they indicate that the proposed geometry is local and direction-family dependent rather than a universal predictor of arbitrary representation changes. The experiments also distinguish mechanistic understanding from optimization. Although complementary representation utility contains statistically detectable predictive structure, neither a learned routing strategy nor several specialized distillation objectives reliably improves upon the CLEAN-KD baseline. Thus, identifying useful local representation structure does not by itself solve the problem of exploiting that structure during global network training.

The main contributions of this work are:
\begin{itemize}
    \item We show that clean-to-degraded task transfer is beneficial on average yet heterogeneous at the object level, with clean-teacher support strongly associated with preservation and rescue.
    
    \item We identify and experimentally intervene on a mapped complementary representation component, showing that it contains downstream-relevant information beyond direct clean-teacher alignment.
    
    \item We introduce a gradient-derived task-sensitive representation geometry and demonstrate prospectively that mapped-complement orientation within this geometry is associated with local intervention response after controlling for intervention magnitude; the relationship replicates on independent data and under a second detector architecture.
    
    \item We characterize the scope of this result through locality, direction-family, routing, and training-objective experiments, showing that representation-level geometric understanding does not automatically yield a universally predictive direction score or a superior global distillation objective.
\end{itemize}

\section{Related Work}
\label{sec:related_work}

\subsection{Object Detection under Image Degradation and Adverse Conditions}
\label{sec:related_degraded_detection}

Visual degradation remains a major challenge for object detection in real-world environments. BDD100K provides diverse driving imagery across geographic, environmental, and weather conditions \citep{Yu2020BDD100K}, while ACDC was introduced specifically to study visual understanding under fog, nighttime, rain, and snow \citep{Sakaridis2021ACDC}. Such datasets have enabled systematic investigation of recognition robustness beyond conventional clear-weather benchmarks.

Approaches to adverse-condition detection include direct training on diverse real-world data, synthetic degradation augmentation, image preprocessing or restoration, and combinations of these strategies. \citet{Gupta2024WeatherDetection}, for example, compared real-weather training, synthetic weather augmentation, and denoising-based preprocessing for object detection and found that visually restoring adverse-weather images did not necessarily provide the strongest downstream detection performance. More recent methods have therefore explored image-adaptive preprocessing optimized in conjunction with detection, such as ERUP-YOLO \citep{Ogino2025ERUPYOLO}. These results reinforce the distinction between improving the visual appearance of an observation and preserving information useful for a detector. Our work addresses a different but complementary level of the problem. Rather than designing an image-enhancement front end, we study how task information is transferred and evaluated within the detector's internal representation under degradation.

\subsection{Task-Driven Image Restoration for Recognition}
\label{sec:related_task_restoration}

Traditional image restoration typically optimizes reconstruction or perceptual criteria, but these criteria need not coincide with downstream recognition utility. This mismatch has motivated task-driven restoration methods that incorporate information from high-level vision models. RestoreX-AI \citep{Marathe2022RestoreXAI} explicitly highlighted cases in which improved apparent image quality did not correspond to improved object-detection behavior. SR4IR \citep{Kim2024SR4IR} introduced a task-driven perceptual loss for super-resolution, using task-network features to encourage restoration of information beneficial to classification, segmentation, and object detection.

Recent work has extended this task-oriented perspective to more general and difficult restoration settings. EDTR \citep{Kim2025EDTR} uses diffusion priors while explicitly preserving task-relevant information under multiple complex degradations, and UniRestore \citep{Chen2025UniRestore} combines perceptual and task-oriented restoration objectives within a unified diffusion-based framework. Collectively, these studies demonstrate that restoration should be evaluated not only by image fidelity but also by its effect on the downstream task.

The present work shares this emphasis on task utility but operates in representation space rather than image space. We do not attempt to restore an image or learn a task-conditioned restoration network. Instead, we ask whether the downstream usefulness of internal feature-space directions can be characterized geometrically and validated through controlled interventions.

\subsection{Knowledge Distillation for Object Detection}
\label{sec:related_kd}

Knowledge distillation transfers information from a teacher model to a student model and was originally developed as a mechanism for transferring the behavior of a stronger or more complex model into a deployable student \citep{Hinton2015KD}. Applying this principle to object detection is more difficult than to image classification because detection combines spatially heterogeneous foreground and background regions, localization and classification objectives, multiscale representations, and large numbers of candidate instances.

A substantial body of work therefore focuses on selecting or structuring the information that should be distilled. Fine-grained feature imitation emphasizes locations near objects rather than uniformly matching full feature maps \citep{Wang2019FineGrainedKD}. Decoupled feature distillation distinguishes foreground and background feature regions and separately treats positive and negative proposals \citep{Guo2021DecoupledFeatures}, while General Instance Distillation transfers feature-, relation-, and response-level knowledge around discriminative instances \citep{Dai2021GID}. Focal and Global Distillation explicitly addresses spatially uneven teacher--student discrepancies through foreground-focused and global relation transfer \citep{Yang2022FGD}. ScaleKD further introduces scale-aware transfer for small-object detection \citep{Zhu2023ScaleKD}.

More recent work continues to question how teacher information should be presented to the student. CrossKD transfers student intermediate features through the teacher detection head to reduce conflicting supervision and provide more task-oriented prediction guidance \citep{Wang2024CrossKD}. KD-DETR addresses the distinct structure of transformer detectors by constructing consistent distillation points \citep{Wang2024KDDETR}. Orthogonal projection has also been investigated as a mechanism for aligning teacher and student representations across tasks and architectures \citep{Miles2024VkD}.

These methods demonstrate that effective distillation depends strongly on representation location, structure, and task context. Our work differs in objective. Rather than introducing another rule for selecting which teacher features should be imitated, we examine the downstream utility of candidate representation displacements themselves. In particular, we test whether a direction's local task response can be related prospectively to its orientation within a gradient-derived task geometry.

\subsection{Gradient-Based Task Sensitivity and Representation Geometry}
\label{sec:related_geometry}

Gradients provide a direct connection between internal representations and downstream model behavior. Gradient-based explanation methods such as Grad-CAM use gradients flowing into convolutional features to identify spatial regions important to a target prediction \citep{Selvaraju2017GradCAM}. Other work has used Fisher-information structure to characterize tasks rather than individual predictions. Task2Vec, for example, constructs task embeddings from Fisher information associated with a probe network and uses those embeddings to compare visual classification tasks and select suitable feature extractors \citep{Achille2019Task2Vec}.

These approaches establish that gradient- or information-derived quantities can reveal task-dependent structure, but their purposes differ from ours. Grad-CAM is primarily an attribution mechanism, whereas Task2Vec constructs a parameter-space embedding for comparing tasks. The proposed task-sensitive geometry is instead defined directly in an internal feature-channel space. It aggregates normalized gradients of the downstream detection loss into a global directional geometry and uses that geometry to score independently constructed representation interventions.

The key methodological distinction is prospective response testing. The geometry is frozen before the intervention outcomes used for confirmation are observed, and its orientation score is evaluated against measured downstream response magnitude while controlling for intervention magnitude. The resulting framework therefore connects task gradients, representation direction, and experimentally measured local task response rather than using gradients solely for visualization or task-level similarity.

\section{Methodology}
\label{sec:methodology}

\subsection{Problem Formulation}
\label{sec:problem}

\subsubsection{Object Detection under Paired Clean and Degraded Observations}
\label{sec:paired}

Let $I_c$ denote a clean image and $I_d$ a degraded observation of the same underlying scene. We consider an object detector trained to operate on $I_d$, while $I_c$ is available during training or controlled analysis but is not required at deployment. This setting permits clean observations to provide task-relevant supervisory information without changing the degraded-image inference setting.

Let $T$ denote a feature representation produced from $I_c$ by a clean-image teacher and $S$ the corresponding representation produced from $I_d$ by a degraded-image student. At a selected feature level, $T,S\in\mathbb{R}^{D\times H\times W}$, where $D$ is the channel dimension and $H\times W$ is the spatial resolution. At spatial location $i$, the corresponding feature vectors are $t_i,s_i\in\mathbb{R}^{D}$.

Our experiments instantiate this formulation at the P4 representation of the YOLO11s detector, for which $D=256$. The formulation itself is independent of this particular dimensionality. A natural clean-teacher displacement at location $i$ is
\begin{equation}
d_{T,i}=t_i-s_i.
\label{eq:teacher_direction}
\end{equation}
Conventional feature-level knowledge transfer encourages the degraded representation to move toward its clean counterpart. This is motivated by the expectation that the clean teacher retains task information that has been weakened or corrupted by image degradation. However, Eq.~\eqref{eq:teacher_direction} also embodies a stronger implicit assumption: that movement toward the clean representation is locally beneficial to the downstream task. This work investigates the limits of this assumption.

\subsubsection{Representation Intervention and Local Task Utility}
\label{sec:local_utility}

We distinguish a representation displacement from its downstream consequence. Let $D_{\mathrm{int}}$ denote an arbitrary feature-map displacement having the same dimensions as $S$. We define a controlled intervention as:
\begin{equation}
S_{D_{\mathrm{int}}}(\alpha)=S+\alpha D_{\mathrm{int}}, \label{eq:intervention}    
\end{equation}
where $\alpha\geq 0$ controls the intervention step size.

Let $L_{\mathrm{det}}(S)$ denote the downstream detection loss evaluated from representation $S$. The task benefit produced by intervention $D_{\mathrm{int}}$ is
$\Delta_{D_{\mathrm{int}}}(\alpha)=L_{\mathrm{det}}(S)-L_{\mathrm{det}}\left(S_{D_{\mathrm{int}}}(\alpha)\right)$.
Accordingly, $\Delta_{D_{\mathrm{int}}}(\alpha)>0$ indicates that the intervention decreases the detection loss, whereas $\Delta_{D_{\mathrm{int}}}(\alpha)<0$ indicates a harmful intervention.

This distinction is central to our formulation. A displacement may be geometrically close to a clean-teacher representation without necessarily producing a favorable task response. Conversely, a direction that is not represented by direct teacher alignment may contain information that benefits the downstream detector.

For comparisons between two intervention families, we additionally define relative transfer utility. If $\Delta_C(\alpha)$ and $\Delta_T(\alpha)$ denote the benefits of complementary and direct teacher interventions, respectively, then
\begin{equation}
Y_{C-T}(\alpha)
=
\Delta_C(\alpha)-\Delta_T(\alpha).
\label{eq:relative_utility}
\end{equation}
Thus, $Y_{C-T}(\alpha)>0$ indicates that, under the specified intervention protocol, the complementary intervention is more beneficial than direct teacher transfer. We use this relative formulation for counterfactual transfer analyses, while the later prospective geometry experiments use the measured response $Y_M$ of the mapped-complement intervention itself. The two quantities are kept distinct throughout the paper.

\subsection{Clean-Task Knowledge Transfer}
\label{sec:clean_transfer}

\subsubsection{Clean-Teacher Distillation}
\label{sec:clean_kd}

We first establish a controlled clean-to-degraded transfer setting. A teacher detector receives the clean image $I_c$, whereas the student receives the paired degraded observation $I_d$. The teacher is used only during training. At inference, the student operates exclusively on degraded input and does not require a clean reference. The student objective combines the standard detection objective with representation-level knowledge distillation:
$\mathcal{L}_{\mathrm{student}}=\mathcal{L}_{\mathrm{det}}+\lambda_{\mathrm{KD}}\mathcal{L}_{\mathrm{KD}}$,
where $\lambda_{\mathrm{KD}}$ controls the relative contribution of clean-teacher supervision. We refer to this training configuration as \emph{CLEAN-KD} and to the corresponding degraded-only training configuration, which does not receive clean-teacher feature supervision, as \emph{NONE}.

The purpose of CLEAN-KD in this work is twofold. First, it provides a practical test of whether clean representations contain information that improves detection from degraded observations. Second, and more importantly for the present study, it creates a controlled representation-transfer setting in which the downstream consequences of clean knowledge can be studied at finer granularity.

\subsubsection{From Aggregate Improvement to Heterogeneous Transfer}
\label{sec:heterogeneous_transfer}

An improvement in aggregate detection accuracy establishes that clean knowledge can be useful on average, but it does not establish that every local transfer is beneficial. Two detectors can differ positively in mean average precision while individual objects undergo qualitatively different transitions. We therefore distinguish object-level transfer outcomes. An object detected both before and after transfer is considered \emph{preserved} (PRES). An object detected by the reference degraded detector but missed after transfer is \emph{lost} (LOST). Conversely, an object missed by the reference detector but recovered after transfer is \emph{rescued} (RESC). Objects missed under both conditions remain \emph{missed} (MISS).

These transitions expose information that aggregate metrics necessarily collapse. In particular, simultaneous occurrence of RESC and LOST objects implies that a globally favorable transfer objective can nevertheless contain locally beneficial and locally harmful representation changes. This observation motivates a more specific question than whether clean knowledge should be transferred: \emph{Which representation directions contain useful task information, and can their utility be characterized before applying the intervention?}

\subsection{Complementary Task Knowledge}
\label{sec:complementary}

\subsubsection{Representation Components beyond Direct Clean-Teacher Alignment}
\label{sec:complement_decomposition}

To investigate whether useful task information exists outside direct clean-teacher alignment, we introduce a degraded representation obtained from an independently trained NONE detector. Let $n_i\in\mathbb{R}^{D}$ denote this representation at spatial location $i$. We decompose $n_i$ relative to the clean-teacher feature $t_i$. Its projection onto $t_i$ and the corresponding residual component are
$e_i=\frac{n_i^\top t_i}{t_i^\top t_i}t_i$ and $c_i=n_i-e_i$.
By construction,
\begin{equation}
c_i^\top t_i=0.
\label{eq:complement_orthogonal}
\end{equation}

We refer to $c_i$ as the \emph{complementary component} in the NONE representation space. Importantly, orthogonality in Eq.~\eqref{eq:complement_orthogonal} does not by itself imply task usefulness. The decomposition is used only to isolate representation content that is not explained by direct alignment with the clean-teacher feature. Whether this complementary component carries useful downstream information is an empirical question tested through controlled interventions.

\subsubsection{Mapping between Independently Learned Representation Spaces}
\label{sec:bridge}

The NONE and CLEAN-KD detectors are trained independently and therefore need not express semantically related features in identical channel coordinates. Directly inserting $c_i$ into the CLEAN-KD representation would consequently confound complementary information with arbitrary coordinate mismatch. We therefore estimate an orthogonal mapping $R$ between the two representation spaces. Given paired representation matrices $N$ (extracted features from NONE) and $S$ (extracted features from CLEAN-KD), the mapping is obtained through the orthogonal Procrustes problem
$R^\star=\arg\min_{R^\top R=I}\left\|NR-S\right\|_F^2$ \citep{Schonemann1966Procrustes}.
For the bridge fit, feature vectors are stacked as rows of $N$ and $S$, matching the implementation convention. The complementary component is therefore mapped into the CLEAN-KD student coordinate system as the row-vector operation
$m_i=c_iR^\star$.
When feature vectors are written as columns elsewhere in the paper, this is the equivalent transpose convention; the underlying orthogonal mapping is unchanged. We refer to $m_i$ as the \emph{mapped complement}. The orthogonality constraint on $R$ is important. It permits a change of representation coordinates while preserving inner products and vector norms under the mapping. The bridge therefore aligns independently learned feature spaces without introducing arbitrary anisotropic scaling that could itself alter the geometric properties subsequently analyzed.

\subsubsection{Controlled Mapped-Complement Intervention}
\label{sec:mapped_intervention}

For an object-level intervention, let $\mathcal{I}$ denote the selected set of P4 support cells. In the frozen mapped-complement rescue experiment, this support is the detector-head support region $R_H(o)$ for ground-truth object $o$, where the subscript $H$ denotes the detector head and indicates that the region is defined on the P4 feature grid used by the detection head rather than directly in image coordinates. Specifically, the exact-box P4 mask contains cells whose centers fall inside the transformed ground-truth box, with a center-cell fallback when no P4 cell center lies inside the box. This mask is then square-dilated by two P4 cells to form $R_H(o)$. Thus, $R_H(o)$ denotes the fixed detector-head support neighborhood over which the object-level intervention is applied; $H$ is a notational label for this head-space support and is not an additional learned quantity or intervention parameter.

We construct a feature-map displacement $D_M$ by placing the mapped-complement vector $m_i$ at each selected cell and zero elsewhere:
$D_M[:,i]=m_i$ for $i\in\mathcal{I}$ and $D_M[:,i]=0$ for $i\notin\mathcal{I}$.
Its total magnitude satisfies the Frobenius-norm relation
$\|D_M\|_F^2=\sum_{i\in\mathcal{I}}\|m_i\|_2^2$.

The controlled mapped-complement intervention is
$S_M(\alpha)=S+\alpha D_M$.
For the frozen LOST/PRES rescue experiment, the mapped complement is applied at the fixed full scale $\alpha=1.0$, with no tuning of intervention scale or support radius. The matched-random control uses the same support and a norm-matched mapped random displacement. This operation provides a direct test of whether the isolated complementary representation structure has measurable downstream consequences. The later prospective task-sensitive-geometry experiments use a separately frozen local intervention protocol with $\alpha=0.25$; the two intervention scales should therefore not be conflated.

\subsection{Task-Sensitive Representation Geometry}
\label{sec:task_geometry}

\subsubsection{From Task Gradients to Directional Sensitivity}
\label{sec:task_gradients}

The intervention experiments establish whether a representation displacement affects the downstream task after it is applied. We next ask whether properties of the displacement can characterize its potential task relevance before observing that response. At spatial location $i$, let
\begin{equation}
g_i
=
\frac{\partial L_{\mathrm{det}}}{\partial s_i},
\qquad
\hat{g}_i
=
\frac{g_i}{\|g_i\|_2},
\label{eq:task_gradient}
\end{equation}
where $g_i\in\mathbb{R}^{D}$ is the gradient of the downstream detection loss with respect to the local feature vector and $\hat{g}_i$ is its normalized direction. The magnitude of $g_i$ can depend on factors such as object difficulty, loss scale, and local confidence. Because our objective is to characterize recurrent \emph{orientations} of task sensitivity rather than allow high-gradient cells to dominate the estimate, the geometry is constructed from the normalized gradients $\hat{g}_i$.

\subsubsection{Global Task-Sensitive Geometry}
\label{sec:global_geometry}

We aggregate normalized gradient directions across the geometry-construction cohort:
\begin{equation}
C_{\mathrm{raw}}
=
\frac{1}{N_g}
\sum_{i=1}^{N_g}
\hat{g}_i\hat{g}_i^\top,
\label{eq:craw}
\end{equation}
where $N_g$ denotes the number of gradient vectors used to construct the geometry. Each outer product $\hat{g}_i\hat{g}_i^\top$ records the channel-space orientation associated with a local task gradient while being invariant to its sign. The aggregation therefore emphasizes directions that repeatedly coincide with task sensitivity across examples and spatial locations.

Although $C_{\mathrm{raw}}$ is symmetric in exact arithmetic, we explicitly symmetrize the empirical matrix to remove possible numerical asymmetry:
$C_{\mathrm{sym}}=\frac{1}{2}\left(C_{\mathrm{raw}}+C_{\mathrm{raw}}^\top\right)$.
We then trace-normalize the matrix:
\begin{equation}
C_g
=
\frac{C_{\mathrm{sym}}}
     {\operatorname{tr}(C_{\mathrm{sym}})}.
\label{eq:cg}
\end{equation}
Thus, $\operatorname{tr}(C_g)=1$. We refer to $C_g$ as the \emph{task-sensitive geometry}.

The interpretation of $C_g$ can be seen by considering a unit direction $u$. Ignoring the numerically negligible symmetrization term for clarity,
\begin{equation}
u^\top C_g u
=
\frac{
\displaystyle
\sum_{i=1}^{N_g}
\left(
u^\top\hat{g}_i
\right)^2
}{
\displaystyle
\sum_{i=1}^{N_g}
\|\hat{g}_i\|_2^2
}.
\label{eq:cg_interpretation}
\end{equation}
Since the nonzero gradients are normalized, $\|\hat{g}_i\|_2=1$, and the denominator in Eq.~\eqref{eq:cg_interpretation} equals $N_g$. The quadratic form therefore measures the average squared alignment between the candidate direction $u$ and the task-gradient orientations represented in the geometry-construction cohort.

A high value indicates that the candidate direction lies preferentially in representation orientations to which the downstream detection task is repeatedly sensitive. Importantly, $C_g$ is not a covariance model of feature appearance and does not describe which feature directions are merely common. It is constructed from gradients of the downstream task and therefore characterizes task sensitivity with respect to representation directions.

\subsubsection{Normalized Task-Sensitive Orientation}
\label{sec:orientation}

For a nonzero candidate displacement $d\in\mathbb{R}^{D}$, we define its normalized task-sensitive orientation as
\begin{equation}
q(d)
=
D
\frac{d^\top C_g d}
     {\|d\|_2^2}.
\label{eq:q}
\end{equation}
The denominator removes the direct effect of displacement magnitude, allowing Eq.~\eqref{eq:q} to characterize orientation rather than intervention size. The factor $D$ provides a convenient isotropic reference. If task sensitivity were distributed uniformly across all channel directions, then $C_g=D^{-1}I$ and $q(d)=1$ for every nonzero $d$. Hence, $q(d)>1$ indicates stronger task-sensitive orientation than the isotropic reference, whereas $q(d)<1$ indicates weaker-than-reference orientation.


\begin{figure}
\centering

\begin{tikzpicture}[
    >=Latex,
    scale=1.0,
    every node/.style={font=\small}
]


\draw[
    rotate=25,
    line width=0.8pt
]
(0,0) ellipse [x radius=3.0cm, y radius=1.15cm];


\foreach \angle in {20,27,34,205,214,222}{
    \draw[
        ->,
        dotted,
        line width=0.6pt
    ]
    (0,0) -- ({2.15*cos(\angle)},{2.15*sin(\angle)});
}


\draw[
    ->,
    line width=1.0pt
]
(0,0) -- (2.7,1.45);

\node[
    anchor=west
]
at (2.72,1.55)
{$d_{\mathrm{high}}$};


\draw[
    ->,
    line width=1.0pt
]
(0,0) -- (-0.55,2.15);

\node[
    anchor=south
]
at (-0.70,1.15)
{$d_{\mathrm{low}}$};


\node[
    rotate=25,
    font=\footnotesize
]
at (1.15,-1.18)
{principal task-sensitive orientation};


\node[
    align=left,
    anchor=west
]
at (-3.5,-2.45)
{
$\displaystyle
q(d)=D\frac{d^\top C_g d}{\|d\|_2^2}
$
\\[1mm]
$q(d)>1$: stronger than isotropic reference
};

\end{tikzpicture}

\caption{
Conceptual illustration of task-sensitive representation geometry.
Normalized downstream task gradients define recurrent task-sensitive
orientations in feature-channel space. Candidate directions of comparable
magnitude can therefore have different orientation scores. A direction such
as $d_{\mathrm{high}}$, which is more strongly aligned with the dominant
task-sensitive orientation, yields a larger $q(d)$ than a direction such as
$d_{\mathrm{low}}$. The illustration is schematic and does not represent an
empirical two-dimensional projection of $C_g$.
}
\label{fig:task_geometry_concept}

\end{figure}
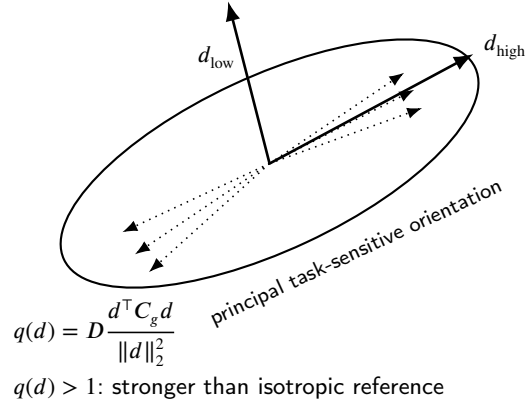

Figure~\ref{fig:task_geometry_concept} illustrates this interpretation geometrically. Because \(C_g\) need not be isotropic, different feature-space directions of comparable magnitude can have substantially different task-sensitive orientation scores. A direction aligned with channel-space orientations that recur in the task gradients receives a larger quadratic-form response \(d^\top C_g d\), and therefore a larger \(q(d)\), than a direction oriented primarily away from those task-sensitive directions. The score thus characterizes directional alignment with recurrent task sensitivity rather than displacement magnitude.

\subsubsection{Multi-Cell Orientation of the Mapped Complement}
\label{sec:multi_cell_orientation}

Mapped-complement interventions generally span multiple spatial cells. Their aggregate orientation is defined as
\begin{equation}
Q_M
=
D
\frac{
\displaystyle
\sum_{i\in\mathcal{I}}
m_i^\top C_gm_i
}{
\displaystyle
\sum_{i\in\mathcal{I}}
\|m_i\|_2^2
}.
\label{eq:qm}
\end{equation}

Equation~\eqref{eq:qm} is an energy-weighted aggregation of the cell-level orientation scores. Cells contributing more strongly to the actual intervention therefore contribute proportionally to its geometric characterization. This is preferable to an unweighted mean of cell-level scores, which would allow arbitrarily small feature vectors to influence the object-level statistic as strongly as dominant components.

A direct comparison between $m_i$ and the gradient $g_i$ at the same spatial cell would answer a different question: whether a displacement aligns with the instantaneous gradient of that particular example. In contrast, $C_g$ summarizes task-sensitive orientations across a separate collection of cells and examples. Equation~\eqref{eq:qm} therefore tests whether an intervention aligns with recurrent global task-sensitive channel geometry rather than with the local gradient from which its own outcome is measured.

\subsubsection{Prospective Orientation--Response Hypothesis}
\label{sec:prospective_hypothesis}

Let $Y_M$ denote the measured downstream response to a mapped-complement intervention. The geometric formulation developed here does not require a particular response metric; the operational definition used in the prospective experiments is specified in Section~\ref{sec:geometry_protocol}. In the primary analysis, $Y_M$ measures the magnitude of the task response induced by a fixed local mapped-complement intervention.

The central hypothesis of this work is that mapped-complement directions more strongly oriented toward the frozen task-sensitive geometry exhibit systematically larger downstream responses ($Q_M$ and $Y_M$ are monotonically correlated). A potential confound is intervention magnitude. Although $Q_M$ is normalized by direction energy, the observed downstream response $Y_M$ can itself depend on $\|D_M\|_F$. We therefore define the primary confirmatory statistic as the partial rank association
\begin{equation}
\rho_{\mathrm{partial}}
\left(
Q_M,
Y_M
\mid
\|D_M\|_F
\right).
\label{eq:primary_partial}
\end{equation}

The hypothesis is evaluated prospectively: $C_g$ is constructed and frozen independently of the intervention outcomes used to evaluate Eq.~\eqref{eq:primary_partial}. Consequently, a positive association cannot arise from fitting the geometry directly to those outcome labels. This prospective separation is important. The objective is not to discover a matrix that retrospectively explains observed intervention effects, but to test whether a task geometry estimated from downstream gradients carries information about subsequently measured local response magnitude.

\subsubsection{Locality of Gradient-Derived Geometry}
\label{sec:locality}

The construction of $C_g$ is fundamentally local because it derives from first-order task gradients. For a sufficiently small feature displacement $D_{\mathrm{int}}$, a local Taylor expansion gives
\begin{equation}
L_{\mathrm{det}}
\left(
S+\alpha D_{\mathrm{int}}
\right)
=
L_{\mathrm{det}}(S)
+
\alpha
\left\langle
\nabla_S L_{\mathrm{det}},
D_{\mathrm{int}}
\right\rangle
+
O(\alpha^2).
\label{eq:taylor}
\end{equation}
Equation~\eqref{eq:taylor} expresses the local change in detection loss induced by a feature-space intervention. Here, $\langle \nabla_S L_{\mathrm{det}},D_{\mathrm{int}}\rangle$ is the directional derivative of the detection loss along the intervention direction $D_{\mathrm{int}}$, with the inner product taken over the feature-map entries. The term $\alpha\langle \nabla_S L_{\mathrm{det}},D_{\mathrm{int}}\rangle$ therefore represents the first-order change in detection loss predicted for a step of size $\alpha$ along that direction. A negative value indicates that a sufficiently small positive intervention is locally expected to reduce the loss, whereas a positive value indicates a locally loss-increasing direction. The remainder term $O(\alpha^2)$ collects second- and higher-order effects, including curvature of the loss surface. Consequently, these higher-order effects become increasingly important as $\alpha$ grows, so a geometry derived from first-order gradients is expected to be most informative for local interventions. This prediction provides a direct test of whether the empirical behavior of $C_g$ is consistent with its gradient-derived interpretation rather than with an unrestricted global linear model of intervention response.

\subsubsection{Direction-Family Generalization}
\label{sec:direction_generalization}

Finally, a positive result for mapped-complement interventions does not imply that $C_g$ universally characterizes the local response magnitude of every representation displacement. To test this boundary, we separately consider the direct clean-teacher residual in Eq.~\eqref{eq:teacher_direction} and construct its corresponding multi-cell orientation score:
\begin{equation}
Q_T
=
D
\frac{
\displaystyle
\sum_{i\in\mathcal{I}}
d_{T,i}^\top C_gd_{T,i}
}{
\displaystyle
\sum_{i\in\mathcal{I}}
\|d_{T,i}\|_2^2
}.
\label{eq:qt}
\end{equation}

For the direct-teacher family, $D_T$ places $d_{T,i}$ at the same frozen support cells $i\in\mathcal{I}$ used by the corresponding mapped-complement test and is zero elsewhere. Its local response magnitude is
$Y_T=\left|\Delta_T(0.25)\right|$.

The mapped-complement and direct-teacher direction families are evaluated separately. Evidence for one is not interpreted as evidence for the other. This distinction defines the intended scope of task-sensitive geometry in the present study. We test whether a frozen gradient-derived geometry characterizes the local response magnitude of a structured mapped-complement direction family and then explicitly test whether that relationship extends to direct clean-teacher residuals. Failure of such extension would constrain, rather than invalidate, the mapped-complement result.

Taken together, Eqs.~\eqref{eq:task_gradient}--\eqref{eq:qt} establish a progression from downstream gradients to a frozen task-sensitive geometry, from geometry to normalized orientation, and from orientation to a prospectively testable hypothesis about representation intervention response.

\section{Experimental Setup}
\label{sec:experimental_protocol}

\subsection{Datasets and Degradation Protocol}
\label{sec:datasets}

The primary experiments use BDD100K with paired clean and synthetically degraded observations. For each source image, Gaussian noise with standard deviations
$\sigma\in\{15,25,50\}$ is applied to generate three degraded views. The clean source image is retained as the paired teacher input. The frozen development split contains 1,500 training sources and 250 validation sources, corresponding to 4,500 and 750 degraded views, respectively.

A separate DEVTEST split contains 250 previously unused source images and is used only after model and method selection. We additionally evaluate CLEAN-KD on ACDC to examine transfer beyond synthetic Gaussian degradation. ACDC is not used for training, hyperparameter selection, representation-geometry construction, or method selection.

BDD100K contains ten detection classes. The primary model-selection endpoint uses the predeclared CORE9 subset, consisting of class IDs
$\{0,1,2,3,4,5,6,7,9\}$ and excluding class 8 (\emph{train}), which has no ground-truth instances in the frozen development validation split and is therefore not imputed. For each Gaussian severity, detection accuracy is measured using mAP$_{50:95}$, and the primary summary metric is the equal-weight mean across the three severities:
$\mathrm{mAP}_{\mathrm{macro}}=\frac{1}{3}\sum_{\sigma\in\{15,25,50\}}\mathrm{mAP}_{50:95}^{(\sigma)}$.

\subsection{Detector Architectures}
\label{sec:architectures}

The principal experiments use YOLO11s \citep{JocherQiu2024YOLO11}. Representation transfer and geometric analyses are performed at the neck P4 feature map, corresponding to module 19 and a tensor of size
$256\times40\times40$ for $640\times640$ input images. Thus, the channel dimension in Sections~\ref{sec:complementary} and~\ref{sec:task_geometry} is $D=256$.

To evaluate whether the principal findings depend on the YOLO architecture, we conduct a separate cross-architecture replication study using RT-DETR-L  \citep{Zhao2024RTDETR}. Its P4 representation used for analysis is module 24, which likewise produces a $256\times40\times40$ feature map. This matched representation dimensionality permits the same conceptual intervention and task-sensitive-geometry framework to be evaluated under a substantially different detection architecture.

YOLO11s models are trained for 50 epochs with image size 640, batch size 16, and random seed 2027. RT-DETR-L uses the same number of epochs, image size, and seed, with batch size 8. Unless explicitly stated otherwise, all compared methods within an architecture use the same frozen data partitions, initialization policy, evaluation protocol, and model-selection endpoint.

\subsection{Clean-Knowledge Distillation}
\label{sec:experimental_kd}

For YOLO11s, CLEAN-KD transfers the paired clean-teacher P4 representation to the degraded-image student using a same-location feature-alignment loss. The final frozen coefficient is
$\lambda_{\mathrm{KD}}^{\mathrm{YOLO}}=17.961793781$.
This coefficient was calibrated before the subsequent mechanistic experiments and is not retuned using DEVTEST, ACDC, or the prospective geometry cohorts. For RT-DETR-L, the corresponding frozen coefficient is
$\lambda_{\mathrm{KD}}^{\mathrm{RT}}=3.826536064$.
The architecture-specific coefficients reflect separate gradient-scale calibration and should therefore not be interpreted as directly comparable measures of knowledge-transfer strength. In all CLEAN-KD experiments, the clean teacher is used only during training. The final student performs inference using only the degraded input.

\subsection{Object-Level Transfer Analysis}
\label{sec:object_analysis}

To characterize transfer below the aggregate mAP level, detections are matched to ground-truth objects using class-aware one-to-one matching with confidence threshold 0.25 and IoU threshold 0.50. The resulting object states are categorized as PRES, LOST, RESC, or MISS as defined in Section~\ref{sec:heterogeneous_transfer}. We additionally classify a ground-truth object according to whether the paired clean teacher provides direct detection support. An object is \emph{teacher-supported} when a post-NMS clean-teacher detection of the same class matches the ground-truth box at confidence at least 0.25 and IoU at least 0.50; otherwise, it is \emph{teacher-unsupported}. These analyses are used as mechanism diagnostics and predictive-association tests rather than as confirmatory tests of a new training method.

\subsection{Frozen Mapped-Complement Intervention Protocol}
\label{sec:mapped_rescue_protocol}

The object-level causal rescue analysis is performed on a frozen validation cohort and modifies only the CLEAN-KD P4 representation within the fixed head-support region $R_H(o)$ defined in Section~\ref{sec:mapped_intervention}. For each target object, both the structured mapped-complement intervention and the specificity control are instances of Eq.~\eqref{eq:intervention}: the former uses $D_{\mathrm{int}}=D_M$, whereas the latter uses a norm-matched random displacement over the same spatial support. The intervention scale is fixed at $\alpha=1.0$, and the head-support dilation radius is fixed at two P4 cells; neither quantity is tuned on validation outcomes. This frozen rescue protocol is distinct from the subsequent task-sensitive-geometry confirmation protocol, for which the local intervention scale is prospectively fixed at $\alpha=0.25$.

\subsection{Geometry Construction, Confirmation, and Replication}
\label{sec:geometry_protocol}

The task-sensitive geometry is constructed independently of the outcomes used for its prospective evaluation. For the primary YOLO11s confirmation, a frozen geometry-construction cohort contains 120 previously unused BDD100K sources evaluated at all three Gaussian severities, giving 360 degraded views. At eligible CORE9 object-center P4 cells, the detection-loss gradients are extracted, normalized according to Eq.~\eqref{eq:task_gradient}, and aggregated into $C_g$ using Eqs.~\eqref{eq:craw}--\eqref{eq:cg}.

A separate reserved cohort of 120 sources is then used to evaluate the orientation--response hypothesis. Neither the mapped-complement intervention responses nor any confirmation-cohort outcome values are used to construct or adapt $C_g$. In this prospective geometry experiment, the mapped-complement intervention scale is fixed at $\alpha=0.25$, independently of the full-scale $\alpha=1.0$ rescue experiment described in Section~\ref{sec:mapped_rescue_protocol}. For each eligible confirmation view, $D_M$ is instantiated simultaneously at all unique CORE9 object-center P4 support cells; no top-fraction or outcome-dependent cell selection is performed. The response variable is the magnitude of the local mapped-complement task response, $Y_M = \left| \Delta_M(0.25) \right|$.
The primary statistic is the partial Spearman correlation in Eq.~\eqref{eq:primary_partial}, controlling for $\|D_M\|_F$.

A subsequent independent fresh-cohort experiment evaluates replication of the mapped-complement orientation result and separately tests whether the same geometric relationship extends to direct clean-teacher residuals. RT-DETR-L provides an additional architecture-level replication using an independently frozen RT-DETR task-sensitive geometry.

\subsection{Statistical Inference}
\label{sec:statistics}

Correlation analyses use Spearman rank correlation because the central hypotheses concern monotonic association rather than a predetermined linear relationship. Partial Spearman correlation is used to control for intervention magnitude when testing orientation--response association. Where confidence intervals are reported, resampling is performed at the source-image level so that the three degradation severities corresponding to one source remain clustered. Confirmatory geometry analyses use 5,000 source-cluster bootstrap replicates and two-sided percentile 95\% confidence intervals unless otherwise stated. We distinguish three levels of statistical interpretation throughout the results. First, \emph{descriptive} results summarize observed effects without a confirmatory claim. Second, \emph{diagnostic} analyses examine mechanisms or associations within the frozen evaluation framework. Third, \emph{confirmatory} results correspond to prospectively specified hypotheses evaluated on reserved data. This distinction is particularly important because confidence intervals were not computed for every earlier DEVTEST or ACDC performance comparison.

\section{Results}
\label{sec:results}

\subsection{Detection Performance and Transfer Heterogeneity}
\label{sec:performance_heterogeneity}

\subsubsection{Clean Knowledge Improves the Observed Detection Accuracy}
\label{sec:clean_performance}

\begin{figure}
    \centering
    \includegraphics[width=0.92\linewidth]
    {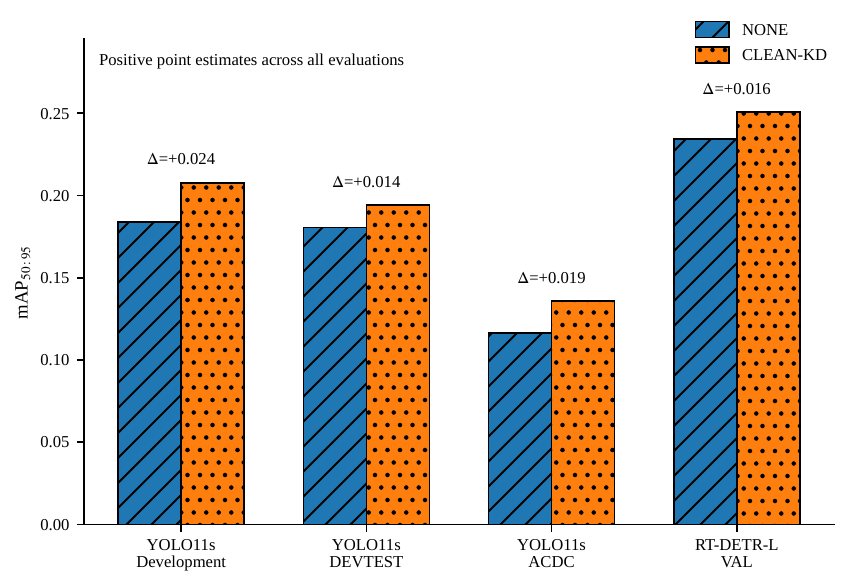}
    \caption{
    Detection performance of degraded-only training (NONE) and
    clean-teacher distillation (CLEAN-KD) across the evaluated settings.
    CLEAN-KD produces positive observed mAP$_{50:95}$ differences on
    YOLO11s Gaussian development, Gaussian DEVTEST, ACDC, and RT-DETR-L
    Gaussian validation. DEVTEST and ACDC are descriptive point
    estimates because paired confidence intervals were not computed for
    these evaluations. For RT-DETR-L, the macro difference is positive,
    but its paired 95\% confidence interval includes zero.
    }
    \label{fig:clean_kd_performance}
\end{figure}

For YOLO11s, CLEAN-KD increases the severity-macro mAP$_{50:95}$ from 0.183919 to 0.207666 on the Gaussian development evaluation, an absolute gain of 0.023747. The same positive direction is observed on the previously unused DEVTEST cohort, where the metric increases from 0.180557 to 0.194197, and on ACDC, where it increases from 0.116531 to 0.135741. Figure~\ref{fig:clean_kd_performance} summarizes these comparisons together with the corresponding RT-DETR-L result. The positive point estimate obtained during YOLO11s development is therefore preserved on fresh synthetic data and on real adverse-condition imagery. Because paired source-bootstrap intervals were not computed for DEVTEST or ACDC, we interpret these results as repeated positive performance observations rather than as separate confirmatory significance tests.

The same directional pattern is observed with RT-DETR-L. Its severity-macro mAP$_{50:95}$ increases from 0.234488 under NONE to 0.250636 under CLEAN-KD, corresponding to an absolute gain of 0.016148. The gain is positive at all three Gaussian severities: $\Delta_{\sigma15}=+0.020303$, $\Delta_{\sigma25}=+0.016249$, and $\Delta_{\sigma50}=+0.011892$. However, the paired source-bootstrap 95\% confidence interval for the RT-DETR-L macro difference is $[-0.004753,+0.038907]$, which includes zero. We therefore interpret the RT-DETR-L result as evidence of cross-architecture directional consistency in the observed performance effect, rather than as statistically established superiority of CLEAN-KD over NONE.

\subsubsection{Clean-Teacher Support Is Associated with Preservation and Rescue}
\label{sec:teacher_support}

The aggregate performance results do not indicate which objects benefit from clean knowledge. We therefore examine object-level transfer conditional on whether the clean teacher itself provides direct support for an object. Teacher-supported objects are substantially more likely to remain detected after clean-knowledge transfer. Their preservation rate is 94.77\%, compared with 53.09\% for teacher-unsupported objects, giving a difference of $+41.68\ \text{percentage points}$, with 95\% confidence interval $[+37.67,\,+45.94]\ \text{percentage points}$.

A similarly large difference occurs for rescue. Among objects eligible for rescue, the rate is 41.69\% for teacher-supported objects and 5.50\% for teacher-unsupported objects, a difference of 
$+36.19\ \text{percentage points}$ with 95\% confidence interval $[+32.43,\,+39.97]\ \text{percentage points}$.

The pattern remains strong for small objects. Teacher support is associated with a preservation advantage of 39.52 percentage points, with 95\% CI $[+34.80,+44.77]$, and a rescue advantage of 32.00 percentage points, with 95\% CI $[+27.26,+36.75]$.

These results indicate that clean-teacher knowledge is strongly associated with object-level preservation and recovery. At the same time, the existence of LOST and teacher-unsupported cases shows that direct clean-teacher alignment does not fully characterize useful representation transfer. This motivates the complementary-knowledge intervention studied next.

\subsection{Controlled Complementary-Knowledge Interventions}
\label{sec:complement_interventions}

\subsubsection{Can the Mapped Complement Rescue Lost Objects?}
\label{sec:lost_rescue}

We next test whether the mapped complementary component defined in Section~\ref{sec:complementary} has a measurable causal effect on downstream detections. The principal analysis focuses on objects that are LOST under the CLEAN-KD representation and compares the structured mapped-complement intervention with a norm-matched random intervention. Among 976 LOST objects, the mapped-complement intervention rescues 129 ( $13.22\%$). Under the matched random intervention, only 42 of 976 are rescued ($4.30\%$).

Thus, the structured intervention produces approximately three times as many rescues as the random perturbation despite matching intervention magnitude. This comparison provides stronger evidence than a representation-similarity statistic alone. The mapped complement is not merely correlated with LOST objects; introducing that structured feature displacement changes the downstream detection outcome more frequently than an energy-matched random direction.

\begin{figure}
    \centering
    \includegraphics[width=0.96\linewidth]
    {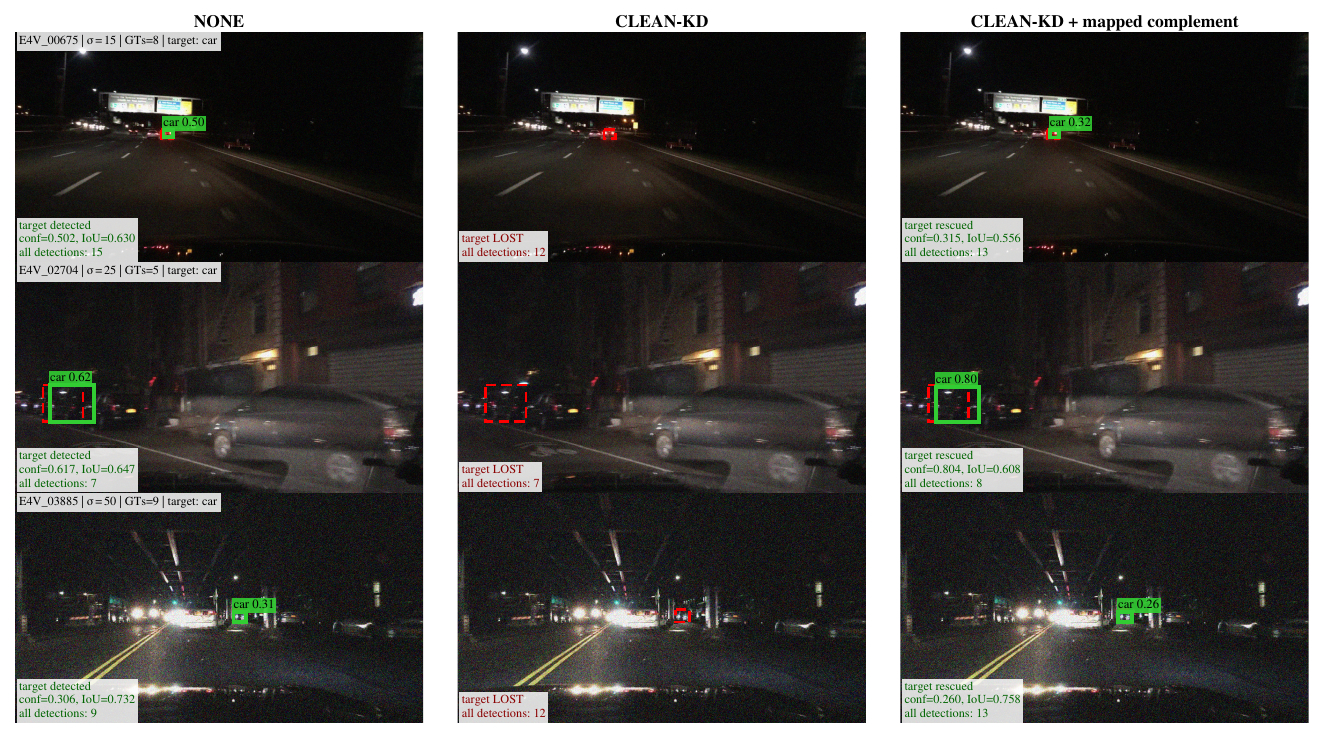}
    \caption{
    Qualitative examples of mapped-complement rescue under Gaussian image
    degradation. Each row shows a frozen validation case in which the target
    object is detected by the degraded-only NONE model, lost under CLEAN-KD,
    and recovered after injecting the basis-mapped complementary
    representation into the frozen CLEAN-KD feature map. One confirmed
    mapped-complement rescue is selected at each Gaussian degradation severity
    $\sigma=15$, $25$, and $50$ using a deterministic fixed-seed procedure
    (seed 2027) from the 129 targeted rescues, with distinct source images.
    Selection does not condition on post-intervention confidence, IoU, or the
    matched-random control outcome. Red dashed boxes indicate the target
    ground-truth object, and green boxes indicate the matched target prediction
    when present. Non-target detections are suppressed for visual clarity.
    These examples illustrate the object-level intervention mechanism and do
    not represent a separately trained detector or a TSG-guided inference
    method.
    }
    \label{fig:qualitative_mapped_rescue}
\end{figure}

The object-level effect underlying this quantitative result is illustrated
in Fig.~\ref{fig:qualitative_mapped_rescue}. Each example shows a target
object that is detected by the degraded-only NONE model, becomes LOST under
CLEAN-KD, and is recovered after the mapped-complement intervention. The
fixed qualitative selection procedure chooses one confirmed rescue per
Gaussian severity without using post-intervention confidence, IoU, or
random-control outcome as a selection criterion. The examples are shown only
to illustrate the intervention mechanism; the statistical comparison with
the norm-matched random control is given by the full cohort analysis above.

\subsubsection{Does Complementary Rescue Damage Preserved Objects?}
\label{sec:preserved_harm}

A useful rescue mechanism should not obtain its apparent benefit by broadly destabilizing objects that were already detected. We therefore perform the corresponding harm analysis on PRES objects. Among 7,115 preserved objects, the mapped-complement intervention harms 33 ($0.464\%$). The matched random intervention harms 37 ($0.520\%$). The mapped-complement intervention therefore increases LOST-object rescue without a corresponding increase in preserved-object harm relative to the random control. The resulting utility contrast is positive,
$\Delta_U=+0.011247$, with 95\% confidence interval $[+0.006925,\,+0.015698]$. These controlled interventions establish that the complementary representation component can contain downstream-relevant information. They do not yet explain why one complementary direction is more useful than another. That question is addressed by the task-sensitive geometry.

\subsection{Prospective Validation of Task-Sensitive Geometry}
\label{sec:prospective_validation}

\subsubsection{Orientation Is Associated with Local Mapped-Complement Response}
\label{sec:orientation_result}

The primary prospective experiment tests whether the frozen orientation score $Q_M$ is associated with the subsequently measured response $Y_M$ on the reserved confirmation cohort. The ordinary rank association is $\rho(Q_M,Y_M)=+0.229754$
with a source-bootstrap 95\% confidence interval of $[+0.103206,\,+0.346938]$. Because larger interventions could potentially induce larger responses regardless of orientation, the predeclared primary analysis controls for $\|D_M\|_F$. The resulting partial Spearman correlation is
\begin{equation}
\rho_{\mathrm{partial}}\left(Q_M,Y_M\mid\|D_M\|_F\right)=+0.241502
\label{eq:primary_geometry_result}
\end{equation}
with 95\% confidence interval $[+0.107726,\,+0.359020]$. The interval lies entirely above zero. Thus, the prospectively frozen task-sensitive geometry contains information about the subsequent local response of mapped-complement interventions even after controlling for intervention magnitude.

\subsubsection{The Association Is Not Explained by Intervention Magnitude}
\label{sec:magnitude_control}

As a diagnostic check, intervention magnitude itself has essentially no monotonic association with the measured response: $\rho\left(\|D_M\|_F,Y_M\right) \approx +0.0055$. This result helps distinguish the geometric effect from a simpler explanation in which larger feature perturbations merely cause larger downstream changes. Together with Eq.~\eqref{eq:primary_geometry_result}, the evidence indicates that the \emph{orientation} of the mapped-complement displacement relative to the frozen task-sensitive geometry carries information beyond its total magnitude.

\subsection{Replication, Generalization, and Locality}
\label{sec:replication_generalization}

\subsubsection{Independent Replication of Mapped-Complement Orientation}
\label{sec:geometry_replication}

We next evaluate the same orientation--response relationship on an independent fresh cohort. Without redefining the scientific hypothesis, the mapped-complement analysis again produces a positive partial rank association:
$\rho_{\mathrm{partial}}\left(Q_M,Y_M\mid\|D_M\|_F\right)=+0.180221$
with 95\% confidence interval $[+0.043453,\,+0.300537]$. Although the point estimate is smaller than in the prospective confirmation experiment, the interval again remains above zero. The relationship between mapped-complement orientation and local response therefore replicates on independent data.

\subsubsection{The Geometry Does Not Generalize to Every Direction Family}
\label{sec:teacher_direction_result}

We then test whether the same geometric relationship extends to the direct clean-teacher residual $d_T=T-S$. Using the corresponding orientation score $Q_T$, the partial association is
$\rho_{\mathrm{partial}}\left(Q_T,Y_T\mid\|D_T\|_F\right)=+0.021715$ with 95\% confidence interval $[-0.118729,\,+0.159348]$. This result does not support direction-family generalization. The positive mapped-complement result should therefore not be interpreted as evidence that $C_g$ is a universal predictor of the local response magnitude of arbitrary feature-space displacements. Instead, the experiments establish a narrower and more informative result: task-sensitive orientation is associated with local response for the structured mapped-complement direction family, and that relationship replicates independently, while extension to a distinct direct-teacher residual family is not established.

\subsubsection{Task-Sensitive Geometry Is Primarily Local}
\label{sec:locality_result}

The Taylor analysis in Section~\ref{sec:locality} predicts that a first-order gradient-derived geometry should be most informative for relatively local feature interventions. We test this expectation by comparing the orientation--response association at a small intervention step, $\alpha=0.25$, and a substantially larger step, $\alpha=4$. The observed Spearman association is
$\rho_{\alpha=0.25}=+0.282684$,
whereas at the larger intervention step it decreases to
$\rho_{\alpha=4}=+0.078321$.
The difference is
$\Delta\rho=+0.204363$
with 95\% confidence interval
$[+0.095329,\,+0.308027]$. We treat this analysis as locality evidence rather than as the primary confirmatory endpoint. Nevertheless, the reduction in association at the larger intervention horizon is consistent with the first-order interpretation of $C_g$: as the representation moves farther from its original operating point, higher-order properties of the detection loss increasingly influence the outcome.

\subsubsection{Replication with RT-DETR-L}
\label{sec:rtdetr_geometry}

\begin{figure}
    \centering
    \includegraphics[width=0.92\linewidth]
    {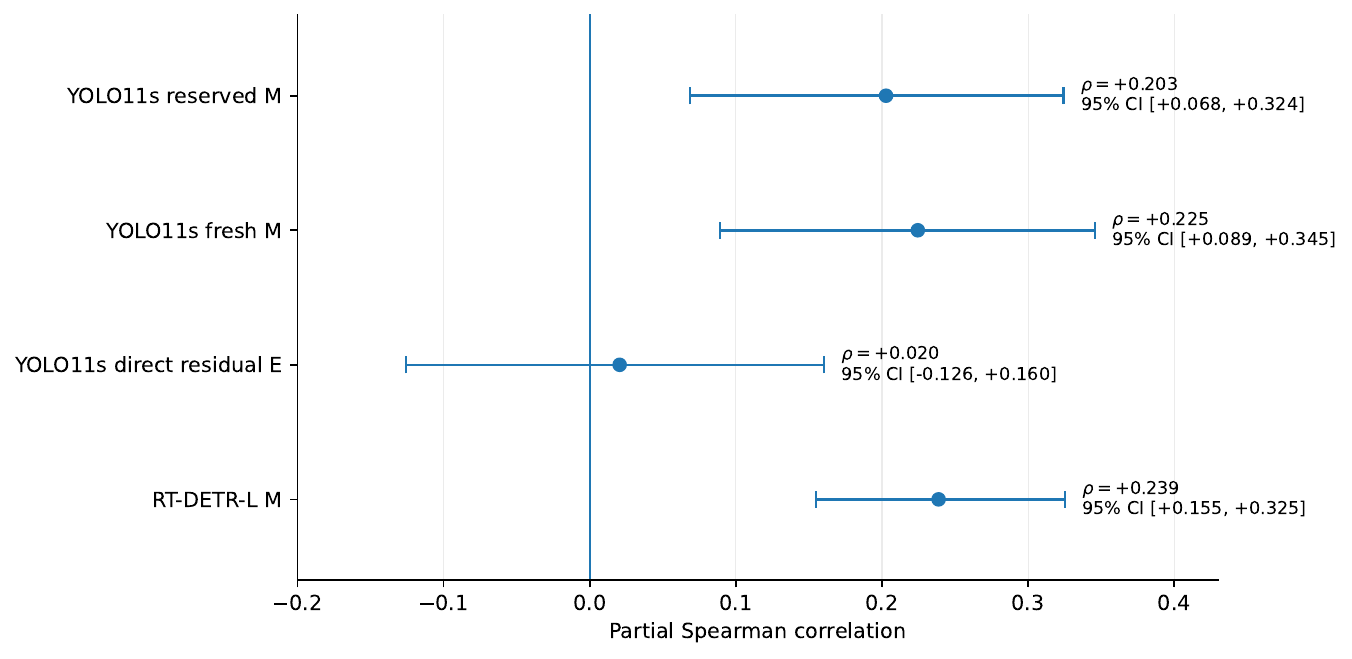}
    \caption{
    Summary of the principal task-sensitive-geometry validation results.
    The prospective mapped-complement analysis shows a positive partial
    association between task-sensitive orientation and local intervention
    response after controlling for intervention magnitude. The relationship
    replicates on an independent fresh cohort and under RT-DETR-L. In
    contrast, the corresponding analysis for the direct clean-teacher
    residual does not support direction-family generalization. The locality
    analysis further shows a stronger orientation--response association for
    the local intervention regime than for the substantially larger
    intervention step. Points denote observed correlation estimates or,
    for the locality analysis, the difference in correlations; horizontal
    bars denote source-bootstrap 95\% confidence intervals.
    }
    \label{fig:geometry_validation}
\end{figure}

The mapped-complement geometry is further evaluated using RT-DETR-L. A separate RT-DETR task-sensitive geometry is constructed from its frozen CLEAN-KD representation and evaluated on a 200-source confirmation cohort containing 600 degraded views across the three Gaussian severities. The primary RT-DETR partial association is
$\rho_{\mathrm{partial}}\left(Q_M,Y_M\mid\|D_M\|_F\right)=+0.226798$
with 95\% source-bootstrap confidence interval
$[+0.146558,\,+0.304902]$.
The corresponding ordinary Spearman correlation is $+0.228888$. Thus, the central mapped-complement orientation relationship is reproduced under a substantially different detection architecture. The absolute strengths of the geometric effects are not assumed to be architecture invariant. Rather, the RT-DETR result supports the broader conclusion that structured complementary directions can exhibit reproducible alignment with downstream task-sensitive geometry across detector families. Figure~\ref{fig:geometry_validation} summarizes the confirmatory, replication, boundary, and locality evidence for task-sensitive geometry across the evaluated analyses.

\subsection{From Geometric Understanding to Trainable Transfer}
\label{sec:geometry_to_training}

\subsubsection{Local Utility Is Predictable but Difficult to Route}
\label{sec:utility_routing}

A natural next question is whether representation-level utility can be predicted accurately enough to select between intervention families automatically. We therefore construct a frozen counterfactual utility predictor using $Y_{C-T}=\Delta_C-\Delta_T$
as defined in Eq.~\eqref{eq:relative_utility}. The predictor is trained only on the training cohort using source-grouped cross-validation, with the validation cohort reserved for evaluation. On the frozen validation set, predicted and observed utility have a positive Spearman correlation:
$\rho(\widehat{Y}_{C-T},Y_{C-T})=+0.03724$
with 95\% confidence interval
$[+0.01128,\,+0.06362]$.
The effect is small but statistically positive, establishing that some representation-level utility information is predictable. However, converting this weak predictive association into a routing policy does not improve utility. Relative to always selecting the direct-teacher intervention, the frozen routing policy produces
$\Delta_{\mathrm{route}}=-0.001058$
with 95\% confidence interval
$[-0.001512,\,-0.000584]$. Thus, predictability alone is insufficient for useful decision routing. A representation signal can contain statistically detectable information while still being too weak, poorly calibrated, or too distribution-dependent to improve a downstream policy.

\subsubsection{Geometry-Inspired Training Objectives Do Not Reliably Surpass CLEAN-KD}
\label{sec:specialized_kd}

We also evaluate whether the mechanistic observations can be translated directly into improved global training objectives. Several prospectively defined variants target different aspects of the observed transfer structure, including boundary-aware transfer, teacher-support-conditioned transfer, complementary transfer, gated complementary transfer, and predictive knowledge distillation.

For the gated complementary objective, we additionally examine the spatial extent of the object-level P4 support. Let $B(o)$ denote the exact-box P4 support for ground-truth object $o$, consisting of P4 cells whose centers lie within the transformed ground-truth box, with the previously defined center-cell fallback when this set is empty. We define
$
R_0(o)=B(o),
\qquad
R_2(o)=\operatorname{Dilate}_{\square}\!\left(B(o),2\right).
$
Thus, R0 denotes the undilated exact-box support, whereas R2 denotes the same support after square dilation by two P4 cells in each spatial direction. The R0--R2 comparison changes only this spatial support extent within the gated complementary objective. Table~\ref{tbl:specialized_methods} summarizes the principal outcomes.

\begin{table}[htbp]
\caption{Summary of specialized transfer objectives. For GCOMP-KD, R0 denotes undilated exact-box P4 support and R2 denotes the same support square-dilated by two P4 cells. Results are not interpreted as a common ranking because several variants address different hypotheses and, where noted, lack a paired confidence interval against CLEAN-KD.}
\label{tbl:specialized_methods}
\begin{tabular*}{\tblwidth}{@{\extracolsep{\fill}}llll@{}}
\toprule
Method & Architecture & Macro mAP$_{50:95}$ & Main conclusion \\
\midrule
CLEAN-KD & YOLO11s & 0.207666 & Reference method \\
BOUNDARY-KD & YOLO11s & 0.205126 & Lower macro point estimate \\
PS-CLEAN-KD (exact-box) & YOLO11s & 0.205226 & Lower macro point estimate \\
COMP-KD & YOLO11s & 0.207522 & No reliable improvement \\
GCOMP-KD (R2) & YOLO11s & 0.210226 & No reliable improvement established \\
GCOMP-KD (R0) & YOLO11s & 0.211951 & R0--R2 CI includes zero \\
PKD & RT-DETR-L & 0.240184 & No reliable improvement over CLEAN-KD \\
CLEAN-KD & RT-DETR-L & 0.250636 & Reference method \\
\bottomrule
\end{tabular*}
\end{table}

The PS-CLEAN-KD entry reports the frozen CORE9 severity-macro endpoint for the prospectively locked exact-box binary-suppression intervention. It should not be confused with a pooled native-evaluation value. The exact-box variant yields a lower macro point estimate than CLEAN-KD, which remains the reference distillation method.

For COMP-KD, the difference from CLEAN-KD is $-0.000143$, with paired 95\% confidence interval $[-0.009521,\allowbreak+0.012592]$. Thus, the complementary objective does not establish a reliable improvement. Within the gated complementary variants, the undilated R0 variant yields a higher point estimate than the radius-two R2 variant at all three severities, but the macro difference is only $+0.001725$, with 95\% confidence interval $[-0.007349,\,+0.009981]$. No reliable directional difference is therefore established between the two support radii. For RT-DETR-L, PKD obtains a macro mAP$_{50:95}$ of 0.240184, compared with 0.234488 for NONE and 0.250636 for CLEAN-KD. Although PKD exceeds NONE by $+0.005696$, its 95\% confidence interval,
$[-0.009907,\,+0.021375]$, includes zero. CLEAN-KD also exceeds PKD by a point estimate of $+0.010452$, but the corresponding interval, $[-0.008303,\,+0.031162]$, likewise includes zero.

Taken together, these experiments reveal an important boundary between mechanistic understanding and global optimization. Controlled interventions show that complementary representation directions can contain useful task information, and the prospective geometry experiments show that local response magnitude is related to task-sensitive orientation. Neither result implies that emphasizing those directions globally during training will necessarily improve detection accuracy. The final evidence therefore supports the distinction: \emph{representation-level understanding} does not imply \emph{successful global optimization}. CLEAN-KD remains the reference distillation method, while task-sensitive geometry serves as a characterization of when structured representation directions are locally relevant rather than as a completed replacement training objective.

\section{Discussion}
\label{sec:discussion}

\subsection{Clean Knowledge Transfer Is Beneficial but Heterogeneous}
\label{sec:discussion_clean_transfer}

The experiments consistently indicate that clean observations contain task-relevant information that can benefit object detection under degradation. With YOLO11s, CLEAN-KD yields higher observed mAP$_{50:95}$ than degraded-only training on the Gaussian development set, the previously unused DEVTEST cohort, and ACDC. RT-DETR-L exhibits the same positive direction across all three Gaussian severities, although its paired confidence interval for the macro improvement includes zero. These results support clean-to-degraded knowledge transfer as a useful experimental foundation, while also showing why aggregate performance alone is insufficient for understanding the transfer process.

The object-level analyses reveal substantial heterogeneity beneath the aggregate improvement. Teacher-supported objects are much more likely to be preserved or rescued, demonstrating that clean-teacher evidence is strongly associated with successful transfer. At the same time, LOST objects remain even under an overall beneficial distillation objective. A representation change that improves the detector globally can therefore still be unfavorable for particular objects.

The coexistence of preservation, rescue, and loss under an overall beneficial transfer objective provides an important indication that clean-task knowledge is not uniformly useful across the representation space. Its downstream value depends on where and how that knowledge is expressed, rather than solely on whether the representation is closer to the clean teacher. This heterogeneity motivates analyzing transfer at the level of representation directions and local task consequences, which is particularly important under image degradation because different objects, spatial regions, and feature channels can be affected unevenly.

\subsection{Complementary Representations Contain Task-Relevant Information}
\label{sec:discussion_complement}

The mapped-complement experiments provide evidence that useful task information is not confined to the representation component directly aligned with the clean teacher. By removing the projection of the independently learned degraded representation onto the clean-teacher feature and mapping the residual into the CLEAN-KD representation space, the analysis isolates a structured direction that is complementary to direct clean-teacher alignment.

Orthogonality alone does not establish usefulness. The stronger evidence comes from intervention. Applying the mapped-complement displacement to LOST objects produces substantially more rescues than an energy-matched random displacement, while the harm rate among PRES objects remains very small and does not exceed the random control. The complementary component therefore has measurable downstream consequences that cannot be explained solely by perturbation magnitude.

This result should not be interpreted as showing that orthogonal representation components are universally preferable to teacher-aligned components. The decomposition provides a controlled mechanism for isolating information outside direct teacher alignment. Its significance is that this isolated structure retains task-relevant information that can affect detection behavior. The experiment therefore demonstrates that conventional movement toward the clean teacher does not exhaust the useful representation directions available to the degraded detector.

\subsection{Task-Sensitive Geometry Characterizes Local Response Magnitude}
\label{sec:discussion_geometry}

The principal contribution of this work is the task-sensitive geometry $C_g$, which summarizes recurrent channel-space orientations of downstream detection sensitivity. Unlike feature covariance, $C_g$ is constructed from normalized gradients of the detection objective and therefore encodes directions to which the task is repeatedly sensitive rather than directions that are simply common in the representation.

The prospective experiment provides the strongest evidence for this interpretation. The geometry is constructed and frozen independently of the reserved intervention outcomes, yet the mapped-complement orientation score $Q_M$ remains positively associated with the subsequently observed response $Y_M$. The association persists after controlling for intervention magnitude, whereas magnitude itself has essentially no monotonic association with the response. The result therefore cannot be reduced to the explanation that larger perturbations simply induce larger effects.

The same mapped-complement orientation relationship replicates on an independent fresh cohort and is reproduced with RT-DETR-L. These results are important because they separate retrospective explanation from prospective characterization. The geometry was not optimized to reproduce the intervention labels on the confirmation sets; instead, task sensitivity extracted from gradients carried information about intervention behavior measured later.

At the same time, the direct clean-teacher residual does not show the same supported orientation--response association. Task-sensitive geometry should therefore not be interpreted as a universal scoring function for arbitrary feature directions. The evidence instead supports a more specific conclusion: within the structured mapped-complement family studied here, representation directions that align more strongly with recurrent task-sensitive orientations tend to produce larger local downstream responses.

\subsection{Why Global Geometry Is Different from a Local Gradient Test}
\label{sec:discussion_global_geometry}

The distinction between $C_g$ and an instantaneous local gradient is important. A direct comparison between an intervention vector $m_i$ and the gradient $g_i$ from the same cell would test whether that intervention aligns with the local first-order loss direction for that particular example. Such a quantity may be useful for explaining an individual intervention, but it would not establish a reusable representation geometry.

In contrast, $C_g$ aggregates normalized task-gradient orientations across many objects, spatial cells, degradation conditions, and images. It therefore asks whether a candidate representation direction aligns with channel-space orientations that repeatedly matter to the downstream task. The prospective evaluation is conducted on separate data, so the geometry is tested as a population-level characterization rather than as an explanation of the same local gradient from which it was constructed.

This distinction also explains why normalization is important. Without normalizing the gradients, a small subset of high-loss or difficult examples could dominate the geometry. The use of normalized gradient directions places the emphasis on recurrent task-sensitive orientation rather than gradient magnitude.

\subsection{The Geometry Is Local and Direction-Family Dependent}
\label{sec:discussion_locality}

Two boundary results help define what the proposed geometry does and does not represent. First, the orientation--response association is substantially stronger for the local intervention step than for the much larger intervention. This behavior is consistent with the first-order construction of $C_g$. A gradient characterizes the local slope of the detection loss around the current representation, whereas sufficiently large movement can enter regions governed by different gradients, nonlinear activations, attention patterns, and higher-order curvature. Task-sensitive geometry should therefore be interpreted primarily as a local representation geometry rather than a global model of the complete detector loss surface. Second, the mapped-complement result does not transfer automatically to the direct clean-teacher residual direction family. This negative generalization result is scientifically useful because it limits the scope of the confirmed claim. The geometry captures task sensitivity and response magnitude, but signed utility also depends on the structure of the candidate intervention family and on how that direction interacts with the representation state. A direction may lie in a task-sensitive subspace yet fail to produce a favorable task effect if its sign, local context, semantic content, or interaction with downstream nonlinearities is inappropriate. Together, these results suggest that signed downstream representation utility cannot be reduced to a single global geometric score. Orientation provides measurable information about response magnitude, but the relationship is local and conditional on the structure of the intervention.

\subsection{Mechanistic Understanding Does Not Automatically Produce a Better Training Objective}
\label{sec:discussion_training}

An important outcome of the study is the separation between representation-level understanding and successful global optimization. The counterfactual utility predictor achieves a statistically positive association with observed relative utility, demonstrating that some information about transfer preference is predictable. Nevertheless, the corresponding routing strategy performs worse than the always-teacher reference. Similarly, the specialized distillation variants motivated by boundary structure, teacher support, complementarity, gating, or predicted utility do not reliably surpass CLEAN-KD.

This apparent discrepancy is not contradictory. A controlled intervention asks what happens when a particular representation direction is changed while most of the detector remains fixed. A training objective changes the representation repeatedly across many examples and optimization steps, thereby altering the distribution on which the original geometric or mechanistic observation was established. Local intervention utility is therefore not equivalent to a globally useful training gradient.

Several additional factors can contribute to this gap. A statistically detectable utility signal may be too weak for reliable per-object decisions; predicted utilities may be imperfectly calibrated; useful intervention directions may vary during training; and emphasizing one locally favorable representation component can introduce interactions elsewhere in the network. The failure of direct translation into a superior global objective is therefore informative rather than merely negative: it identifies the remaining challenge as one of closed-loop optimization, not simply representation discovery. CLEAN-KD consequently remains the reference distillation method in the present study. Task-sensitive geometry is best viewed as a mechanism for understanding and characterizing local task sensitivity and response magnitude, rather than as a completed replacement for conventional knowledge distillation.

\subsection{Implications for Robust Visual Recognition}
\label{sec:discussion_implications}

The results suggest a broader view of robust visual learning under degradation. Image degradation does not merely reduce visual quality; it changes which representation directions are available, reliable, or useful for a downstream task. Consequently, transferring a globally preferred clean representation may not be sufficient when useful information is distributed heterogeneously across feature space.

A task-sensitive approach instead considers three components jointly: the representation direction being introduced, the downstream geometry to which the task is sensitive, and the local state from which the intervention is applied. This view may be useful beyond Gaussian noise or object detection. Restoration, domain adaptation, multimodal transfer, and other settings in which an auxiliary source provides nominally ``better'' information may similarly benefit from distinguishing source fidelity from downstream task utility.

The present results also suggest that future transfer mechanisms may need to be adaptive rather than globally fixed. A successful method may require identifying candidate directions, evaluating their task-sensitive orientation locally, and updating the representation policy as the detector itself changes during training. Such a closed-loop formulation would more directly connect the mechanistic evidence established here with end-to-end optimization.

\section{Limitations and Future Work}
\label{sec:limitations}

Several limitations define the scope of the present conclusions.

First, the primary discovery and confirmatory experiments use Gaussian noise on BDD100K. Evaluation on ACDC provides evidence that the CLEAN-KD model retains a positive performance direction under real adverse conditions, but the full task-sensitive-geometry analysis has not yet been repeated across a broad range of natural degradations such as rain, fog, motion blur, compression artifacts, or low illumination. Establishing whether similar geometric relationships persist across degradation mechanisms is an important next step.

Second, the principal representation analysis is performed at the P4 feature level. P4 provides a useful balance between spatial and semantic resolution and permits matched $256\times40\times40$ representations across the two evaluated architectures, but the results do not establish that the same geometry holds at P3, P5, backbone representations, or decoder features. A multiscale analysis could determine whether task-sensitive orientation differs systematically with representation depth and object scale.

Third, the mapped complement represents one structured intervention family rather than an exhaustive decomposition of feature space. Its prospective orientation relationship replicates, whereas the direct teacher-residual family does not show supported generalization. Other families---including learned basis directions, semantic subspaces, curvature-aware directions, or directions derived from multiple teachers---remain unexplored.

Fourth, $C_g$ is constructed from first-order gradients and is therefore inherently local. The locality experiment supports this interpretation. Modeling intervention response over larger representation displacements may require higher-order information, local recomputation of the geometry, trajectory-dependent analysis, or explicitly nonlinear models of the loss surface.

Fifth, the cross-architecture performance evidence should be interpreted cautiously. RT-DETR-L reproduces the mapped-complement task-sensitive orientation relationship, but the confidence interval for its CLEAN-KD performance difference relative to NONE includes zero. The architecture experiment therefore strengthens the mechanistic generalization of the geometry more clearly than it establishes universal performance gains from clean knowledge distillation.

Sixth, none of the specialized training variants tested here reliably improves upon CLEAN-KD. This limits any claim that the proposed geometric understanding already yields a superior training algorithm. Future work should investigate adaptive objectives that recompute task-sensitive structure during optimization, uncertainty-aware utility estimation, and policies trained directly for downstream improvement rather than for correlation with local intervention outcomes.

Finally, the current experiments focus on object detection. Whether analogous task-sensitive geometries arise in segmentation, tracking, pose estimation, or other structured visual tasks remains unknown. Extending the framework to these settings would help determine which aspects of the observed geometry are detector-specific and which reflect a more general property of task-conditioned representation transfer.

\section{Conclusion}
\label{sec:conclusion}

This work examined clean-to-degraded knowledge transfer from the perspective of downstream representation utility. Clean-teacher distillation yields consistently positive observed detection-performance differences across the primary YOLO11 evaluations, while object-level analysis shows that the underlying transfer is strongly heterogeneous. Clean-teacher support is associated with substantially higher preservation and rescue rates, but aggregate improvement does not eliminate locally unfavorable transfer.

Controlled interventions further show that representation information complementary to direct clean-teacher alignment can affect downstream detections. Mapped-complement interventions rescue LOST objects substantially more often than norm-matched random perturbations while maintaining a very low harm rate among PRES objects. These results establish that useful degraded-image information can exist outside direct teacher-aligned representation components.

To characterize the local task sensitivity of these representation directions, we introduced task-sensitive geometry, a gradient-derived representation of recurrent downstream sensitivity in feature-channel space. The orientation of mapped-complement interventions within this geometry is prospectively associated with their local task-response magnitude after controlling for intervention magnitude. The relationship replicates on independent data and under RT-DETR-L, but does not generalize to the direct clean-teacher residual family and weakens for large interventions. The evidence therefore supports a local, direction-family-dependent response characterization rather than a universal geometric utility model.

Finally, the experiments show that mechanistic understanding does not automatically translate into superior global training. Although representation utility contains detectable predictive structure, routing and specialized distillation objectives do not reliably improve upon CLEAN-KD. The remaining challenge is therefore not only to identify task-relevant representation directions, but to integrate such information adaptively into optimization without introducing negative transfer.

These findings shift the problem from determining whether clean knowledge should be transferred to understanding \emph{which representation directions carry useful task information, when they are useful, and how they can be exploited without inducing negative transfer}.

\printcredits

\bibliographystyle{cas-model2-names}

\bibliography{cas-refs}



\end{document}